\documentclass[a4paper]{article}

\usepackage{algorithm, algorithmic}
\usepackage{amsmath, amssymb, mathtools, bm}
\usepackage{color}

\usepackage[bottom]{footmisc}

\usepackage[utf8]{inputenc}

\usepackage[]{hyperref}
\hypersetup{%
	colorlinks,
	linkcolor={blue},
	citecolor={blue},
	urlcolor={blue}
}

\newcommand{\ket}[1]{\left| #1 \right\rangle}

\newcommand{\rgbsquare}[1]{
	\definecolor{input-color}{RGB}{#1}
	\textcolor{input-color}{\rule[-2pt]{8pt}{8pt}}
}
\newcommand{\whitesquare}{
	\fbox{\tiny\color{white}\tt a}\rule[-5pt]{0pt}{5pt}
}

\newcommand{\rgbsample}[1]{
	\definecolor{input-color}{RGB}{#1}
	\textcolor{input-color}{\rule[-2pt]{8pt}{8pt}}
	(#1)
}

\title{Multi-sensory Integration in a Quantum-Like Robot Perception Model}

\author{%
	Davide Lanza%
	\thanks{%
		Department of Informatics, Bioengineering, Robotics, and Systems Engineering, University of Genoa, Via All'Opera Pia 13, 16145, Genoa, Italy.
		{\tt\footnotesize davide.lanza@eleves.ec-nantes.fr},
		{\tt\footnotesize fulvio.mastrogiovanni@unige.it}}
	\and Paolo Solinas%
	\thanks{%
		Department of Physics, University of Genoa, and National Institute for Nuclear Physics (Genoa section), Via Dodecaneso 33, 16146, Genoa, Italy.
		{\tt\footnotesize solinas@fisica.unige.it}}
	\and Fulvio Mastrogiovanni\footnotemark[1]
}

\date{}

\begin{document}

\maketitle

\begin{abstract}
	Formalisms inspired by Quantum theory have been used in Cognitive Science for decades. Indeed, Quantum-Like (QL) approaches provide descriptive features that are inherently suitable for perception, cognition, and decision processing. A preliminary study on the feasibility of a QL robot perception model has been carried out for a robot with limited sensing capabilities \cite{lanza_preliminary_2020}. 
    In this paper, we generalize such a model for multi-sensory inputs, creating a multidimensional world representation directly based on sensor readings. Given a 3-dimensional case study, we highlight how this model provides a compact and elegant representation, embodying features that are extremely useful for modeling uncertainty and decision. Moreover, the model enables to naturally define query operators to inspect any world state, which answers quantifies the robot's degree of belief on that state.
\end{abstract}

\section{Related Work}
\label{sec:stateofart}

    Quantum computing has been applied in robotics as a tool for speed up classical tools, or in a framework which is still the classical one \cite{petschnigg_quantum_2019}. 
    Our approach is quite different: starting from the very properties of quantum systems, we studied how to exploit them in a novel, simpler framework. 
    This approach has been shown effective in quantum perception and cognition modeling, and we argue that this could be extremely useful in Robotics, providing also a way to translate the theoretical models of quantum cognition to practical robotics application. In contrast to the aforementioned research in Quantum Robotics, this approach could be useful even as purely simulated, because its merits are due to quantum system properties rather than merely computational advantages.

    Since the early intuitions by Amann \cite{amann_gestalt_1993}, quantum cognition research studied the links between perception and quantum dynamics \cite{conte_testing_2008, conte_mental_2009, manousakis_quantum_2009, paraan_more_2014}. A relevant example is the work in Manousakis \cite{manousakis_quantum_2009}, which proposed a QL model to describe probability distributions of perceptive dominances in subjects experiencing binocular rivalry.
    
    A preliminary model inspired by the work in Manousakis \cite{manousakis_quantum_2009} has been proposed to assess the feasibility of a QL perception model for a robot with limited sensing capabilities \cite{lanza_preliminary_2020}. 
    The reason behind this choice is the great descriptive potential which a quantum formalism inherently provides \cite{khrennikov_ubiquitous_2010, busemeyer_quantum_2012, asano_quantum_2015, haven_palgrave_2017, conte_algebraic_2018}.

    Indeed, following Caves \textit{et al.}\cite{caves_quantum_2002}, quantum probability theory can be understood within the Bayesian approach, with probabilities quantifying the degree of belief about a certain state. In this case, maximal information for a question does not imply complete knowledge, i.e., it does not allow us to predict which state will be measured (which answer will be given) but provides only each state's probability of being measured (the degree of belief about the possible answers).
    
    This interpretation of the measurement as a \textit{query} given a certain belief (i.e., the quantum system state) can be extremely useful for decision making \cite{busemeyer_what_2015}. 
    
    We argue that this interpretation could be adapted with significant results in robot perception and cognition models as well. A QL model provides a way to deal with uncertain perceptual knowledge and decision making without an explicit representation. We posit that this is a more elegant, more compact approach because the \textit{state} is not a mere vector state as in current thinking. Representing it as a quantum state, we may be capable of leveraging the properties related to measurement and uncertainty in quantum mechanics. Moreover, this approach discloses new perspectives for further investigations. Starting from a QL representation, a wide range of quantum cognition models discussed in the existing literature can be applied to Robotics \cite{busemeyer_quantum_2012, asano_quantum_2015}.

	The preliminary QL perception model proposed in \cite{lanza_preliminary_2020} dealt with one sensory input channel, performing a time integration of the input discriminating between two states. 
    The goal of this study is to generalize such first single-qubit model to a multi-qubit approach. Moreover, exploiting state superposition and a change of basis in the Hilbert space $H$, we can significantly extend the considered states range. Indeed, we can virtually deal with any possible state in $H$\footnote{
        It is noteworthy that only one query at a time is possible due to the quantum state collapse after any measurement \cite{nielsen_quantum_2010}.
    }. To keep the analysis simpler, we do not consider time integration in this paper, although the model may be easily extended to time windows $\Delta t$ as illustrated for the single-qubit model in \cite{lanza_preliminary_2020}.

\section{Technical Approach}
\label{sec:model}
    
	\begin{figure}[t]
    	\centering
    	\includegraphics[height=4cm]{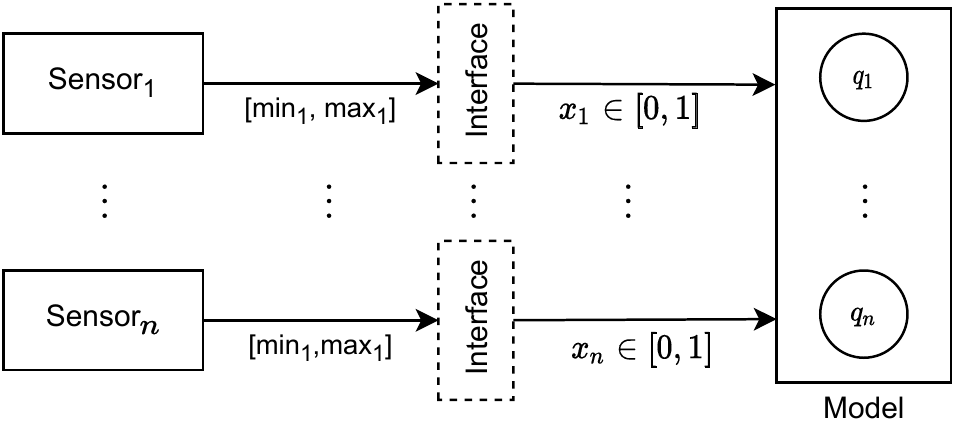}
    	\caption{Sensor and input interfaces scheme for a model made by $n$ qubits $q$.}
    	\label{fig:sensorinterface}
    \end{figure}
    
    \begin{figure}[t]
    	\centering
    	\includegraphics[height=4cm]{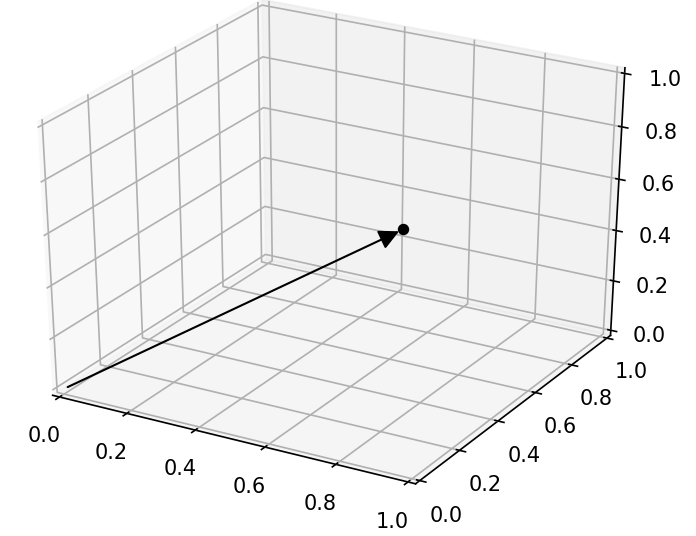}
    	\hspace{1.5cm}
    	\includegraphics[height=4cm]{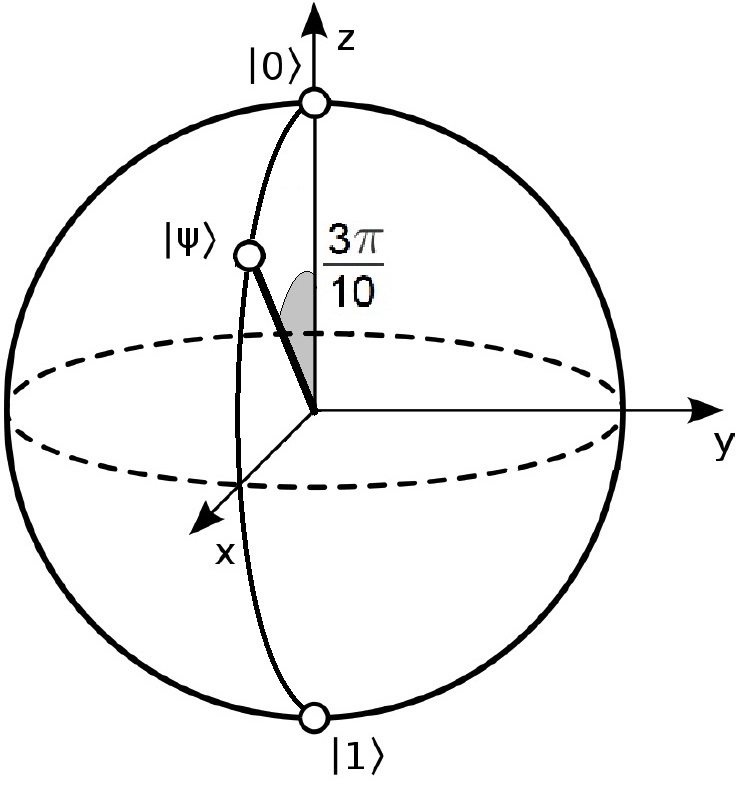}
    	\caption{
            Three-dimensional representation of a vector input $\bm{x}=[0.8,0.3,0.7]$ for a $n=3$ model (left) and the corresponding Bloch sphere representation of the second qubit $q_2$  encoding $x_2 = 0.3$ (right).
        }
        \label{fig:vectorexample}
    \end{figure}    

    We consider $n$ sensors, each one returning a lower- and upper-bounded discrete scalar (Fig. \ref{fig:sensorinterface}). Readings are domain-wise normalized, so the model receives a real value $x_i$ between 0 and 1 for each $i$-th sensor. At each reading update, the model has a vector $\bm{x} = [x_1, ... ,x_n]$ in input. For example, considering a camera-like sensor able to provide only the three RGB average values of the image, we can decompose it in three sensors, each of them with an interface normalizing the $[0,255]\in\mathbb{N}$ readings in a $[0,1]\in\mathbb{R}$ interval. In this case, the input vector $\bm{x}$ is represented in three dimensions as shown in Fig. \ref{fig:vectorexample}.
    Every qubit $q_i$ encodes the sensory information of a corresponding $i$-th sensor. Following \cite{lanza_preliminary_2020} we encode sensory data with a unitary operator $U$ which applies a rotation $R_y\left( \pi x_i / \tau \right)$ to $q_i, \forall i \in [1,n]$. The main differences with the previous model rely on the multi-qubit generalization, the lack of temporal integration ($\tau = 1$ for the model proposed here), and the extension to continuous inputs (the previous study assumed $x_i$ being either 0 or 1). Therefore, the information $x_i$ is encoded in the angle of the Bloch sphere representation of $q_i$, as shown in Fig. \ref{fig:vectorexample}.
    
    Many indirect methods are available to exploit information encoded in qubits \cite{hangos_state_2011}. Here, we consider only state measurements exploiting Caves \textit{et al.} \cite{caves_quantum_2002} interpretation, as stated in Section \ref{sec:stateofart}. Measuring the quantum system leads to the collapse of its state in one of its basis states, namely the set of $2^n$ states composed by all the ordered combinations of the $\ket{0},\ket{1}$ basis states of each single qubit. The probability for the collapse to produce a certain state as a measurement is given by the current state superposition. For example, considering the input vector $\bm x = [0.8, 0.3, 0.7]$ we saw in Fig. \ref{fig:vectorexample}, the overall system state $\psi$ produced by the application of the rotation operators is 
    \begin{equation}
        \ket{\psi} = \left[0.125, 0.385, 0.064, 0.196, 0.245, 0.755, 0.125, 0.385\right].
    \end{equation}

\begin{figure}[t]
	\centering
	\includegraphics[height=5cm]{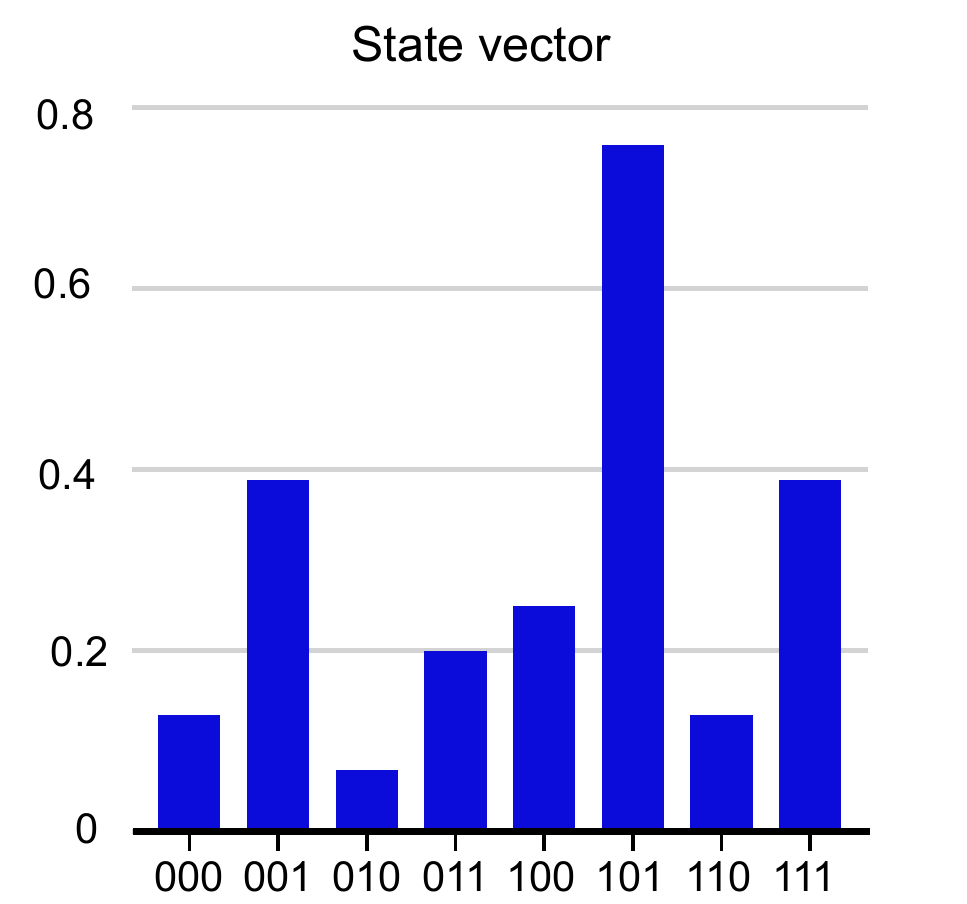}
	\hspace{.5cm}
	\includegraphics[height=5cm]{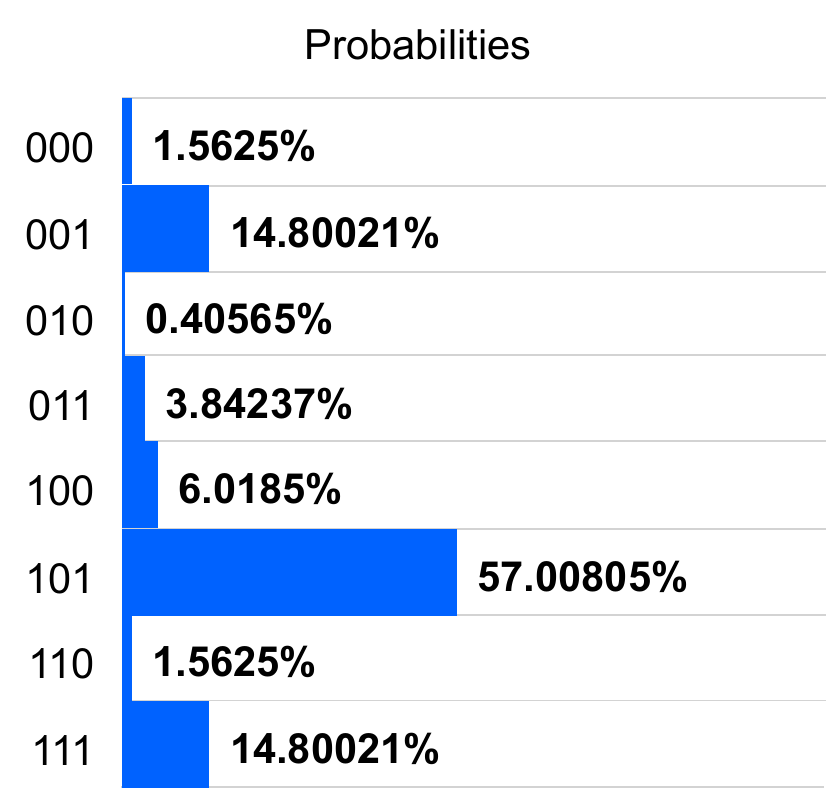}
	\caption{
		The state vector $\ket{\psi}$ histogram and the corresponding measurement probabilities of the quantum model encoding the vector input $\bm{x}=[0.8,0.3,0.7]$ of Figure \ref{fig:vectorexample} (IBM Quantum Experience Circuit Composer simulation).
	}
	\label{fig:stateexample}
\end{figure}

    As illustrated in Fig. \ref{fig:stateexample}, the probability of measuring a certain state is the square of the vector coefficient corresponding to that state. For example, for the $6$-th state $\ket{101}$\footnote{
        We use the IBMQ Qiskit \cite{ibm_quantum_experience_qiskit_2020} notation rather than the one usually used in quantum mechanics. This means that, in the Dirac notation, the states which are normally ordered as $\ket{q_1,\ldots,q_n}$ are instead ordered as $\ket{q_n,\ldots,q_1}$, following the usual bits notation $\ket{\text{MSB},...,\text{LSB}}$.
    }
    we have a coefficient $c_6 = 0.755$ and then the probability of measuring it is $|c_6|^2 = 0.57$ \cite{nielsen_quantum_2010}. These coefficients are related to the input vector due to the applied rotation operators \cite{lanza_preliminary_2020}.
    Since every superposition state is itself a state in the Hilbert space and has a physical meaning on its own, we can operate a basis change. This allows us to define a ``query operator'' $Q(\bar{\bm x})$ which addresses a specific target perceived state $\bar{\bm x}$, applying an inverse rotation $R_y(-\bar{x_i}\pi)$ to $q_i, \forall i \in [1,n]$. This enables us to directly associate the state $\ket{000}$ with this target state. 
    However, by changing the basis we lose the correspondence of the other states with the related sensors. Nevertheless, if we measure the state after applying $Q$, we know that obtaining a state containing ``two zeros'' means measuring a state closer to the target rather then one with just ``one zero'', while $\ket{111}$ is the opposite state. 

\section{Experiments and Results}
\label{sec:test}

We defined a case study considering an ideal camera-like sensor providing the average RGB values of each recorded image. The image has been conceptually decomposed in three scalar sensors (Fig. \ref{fig:sensorinterface}). RGB scalar values can be though of as scalar readings coming from different sensors, even based on different physical transduction mechanisms. However, for representation' sake, we opted for the RGB average values because they allow for a compact, color-related vector representation.

\begin{table}[t]
	\centering
	\resizebox{\linewidth}{!}{%
		\begin{tabular}{|l|l|l|l|l|l|l|l|l|l|r|}
			\hline
			\multicolumn{1}{|c|}{Input (R,G,B)} &
			\multicolumn{1}{c|}{Target (R,G,B)} & 
			\multicolumn{1}{c|}{$\ket{000}$} &
			\multicolumn{1}{c|}{$\ket{001}$} &
			\multicolumn{1}{c|}{$\ket{010}$} &
			\multicolumn{1}{c|}{$\ket{011}$} &
			\multicolumn{1}{c|}{$\ket{100}$} &
			\multicolumn{1}{c|}{$\ket{101}$} &
			\multicolumn{1}{c|}{$\ket{110}$} &
			\multicolumn{1}{c|}{$\ket{111}$} &
			\multicolumn{1}{c|}{$d$}
			\\ \hline
			\multicolumn{11}{c}{}\\[-8pt]
			\hline
			\rgbsample{0,25,0} &
			\multicolumn{1}{c|}{--} &
			\rgbsquare{  0,  0,  0}97.67\% &
			\rgbsquare{255,  0,  0}0\% &
			\rgbsquare{  0,255,  0}2.33\% &
			\rgbsquare{255,255,  0}0\% &
			\rgbsquare{  0,  0,255}0\% &
			\rgbsquare{255,  0,255}0\% &
			\rgbsquare{  0,255,255}0\% &
			\whitesquare 0\% 
			&\multicolumn{1}{c|}{--}
			\\ 
			\rgbsample{55,0,210} &
			\multicolumn{1}{c|}{--} &
			\rgbsquare{  0,  0,  0}6.67\% &
			\rgbsquare{255,  0,  0}8.31\% &
			\rgbsquare{  0,255,  0}0\% &
			\rgbsquare{255,255,  0}0\% &
			\rgbsquare{  0,  0,255}82.27\% &
			\rgbsquare{255,  0,255}10.23\% &
			\rgbsquare{  0,255,255}0\% &
			\whitesquare		   0\% 
			&\multicolumn{1}{c|}{--}
			\\ 
			\rgbsample{10,75,125} &
			\multicolumn{1}{c|}{--} &
			\rgbsquare{  0,  0,  0}41.10\% &
			\rgbsquare{255,  0,  0}0.15\% &
			\rgbsquare{  0,255,  0}10.22\% &
			\rgbsquare{255,255,  0}0.04\% &
			\rgbsquare{  0,  0,255}38.75\% &
			\rgbsquare{255,  0,255}0.14\% &
			\rgbsquare{  0,255,255}9.57\% &
			\whitesquare		   0.04\% 
			&\multicolumn{1}{c|}{--}
			\\ 
			\rgbsample{0,200,200} &
			\multicolumn{1}{c|}{--} &
			\rgbsquare{0,0,0}1.22\% &
			\rgbsquare{255,0,0}0\% &
			\rgbsquare{0,255,0}9.83\% &
			\rgbsquare{255,255,0}0\% &
			\rgbsquare{0,0,255}9.82\% &
			\rgbsquare{255,0,255}0\% &
			\rgbsquare{0,255,255}79.13\% &
			\whitesquare 0\% 
			&\multicolumn{1}{c|}{--}
			\\
			\rgbsample{230,15,230} &
			\multicolumn{1}{c|}{--} &
			\rgbsquare{  0,  0,  0}0.05\% &
			\rgbsquare{255,  0,  0}2.28\% &
			\rgbsquare{  0,255,  0}0\% &
			\rgbsquare{255,255,  0}0.02\% &
			\rgbsquare{  0,  0,255}2.28\% &
			\rgbsquare{255,  0,255}94.55\% &
			\rgbsquare{  0,255,255}0.02\% &
			\whitesquare		   0.81\% 
			&\multicolumn{1}{c|}{--}
			\\ 
			\rgbsample{215,225,220} &
			\multicolumn{1}{c|}{--} &
			\rgbsquare{  0,  0,  0}0.01\% &
			\rgbsquare{255,  0,  0}0.14\% &
			\rgbsquare{  0,255,  0}0.26\% &
			\rgbsquare{255,255,  0}4.15\% &
			\rgbsquare{  0,  0,255}0.19\% &
			\rgbsquare{255,  0,255}3.02\% &
			\rgbsquare{  0,255,255}5.47\% &
			\whitesquare		   86.77 \% 
			&\multicolumn{1}{c|}{--}
			\\ 
			\hline
			\multicolumn{11}{c}{}\\[-8pt]
			\hline
			\rgbsample{102,18,124} &
			\rgbsample{132,35,107} &
			\rgbsquare{132,35,107}94.5\% &
			\rgbsquare{255,255,255}3.3\% &
			\rgbsquare{255,255,255}1.0\% &
			\rgbsquare{255,255,255}0\% &
			\rgbsquare{255,255,255}1.0\% &
			\rgbsquare{255,255,255}0\% &
			\rgbsquare{255,255,255}0\% &
			\rgbsquare{255,255,255}0\% &
			38.44 
			\\
			\rgbsample{84,48,38} &
			\rgbsample{132,35,107} &
			\rgbsquare{132,35,107}75.5\% &
			\rgbsquare{255,255,255}7\% &
			\rgbsquare{255,255,255}0.5\% &
			\rgbsquare{255,255,255}0\% &
			\rgbsquare{255,255,255}15.5\% &
			\rgbsquare{255,255,255}1.4\% &
			\rgbsquare{255,255,255}0.1\% &
			\rgbsquare{255,255,255}0\% &
			85.05 
			\\ 
			\rgbsample{36,101,84} &
			\rgbsample{132,35,107} &
			\rgbsquare{132,35,107}57.0\% &
			\rgbsquare{255,255,255}25.7\% &
			\rgbsquare{255,255,255}10.6\% &
			\rgbsquare{255,255,255}4.8\% &
			\rgbsquare{255,255,255}1.2\% &
			\rgbsquare{255,255,255}0.5\% &
			\rgbsquare{255,255,255}0.2\% &
			\rgbsquare{255,255,255}0\% &
			118.75
			\\ 
			\rgbsample{239, 239,110} &
			\rgbsample{132,35,107} &
			\rgbsquare{132,35,107}6.0\% &
			\rgbsquare{255,255,255}3.6\% &
			\rgbsquare{255,255,255}56.6\% &
			\rgbsquare{255,255,255}33.8\% &
			\rgbsquare{255,255,255}0\% &
			\rgbsquare{255,255,255}0\% &
			\rgbsquare{255,255,255}0\% &
			\rgbsquare{255,255,255}0\% &
			230.38
			\\ 
			%
			\hline
		\end{tabular}
	}
	\caption{The Model behavior in the canonical basis and applying $Q(\bar{\bm x})$, $N=10^6$. For the examples in which a query has been applied, the target and the Euclidean distance $d$(input, target) are reported.}
	\label{tab:colors}
\end{table}

\begin{figure}[b]
	\centering
	\includegraphics[width=\linewidth]{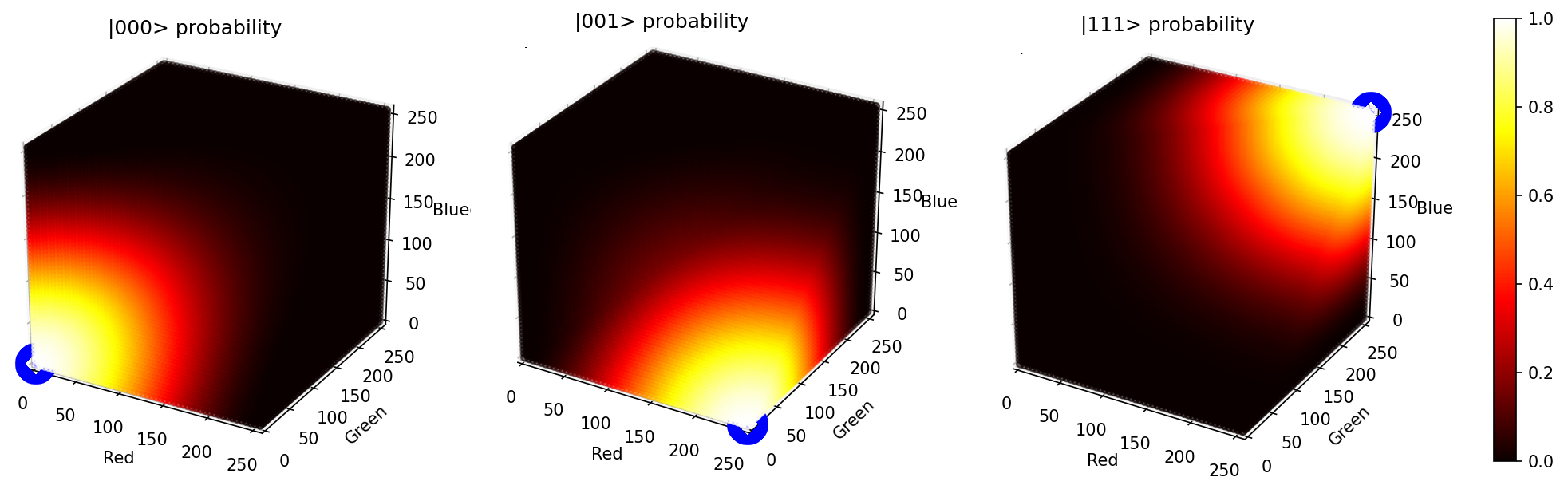}
	\caption{Probabilities spanned through the whole RGB space. Input $\bar{\bm x}$ are points in the RGB cube sampled with a step of 5. No $Q$ applied. The basis state considered is circled in blue.}
	\label{fig:cubestates}
\end{figure}

\begin{figure}[t]
	\centering
	\includegraphics[width=9cm]{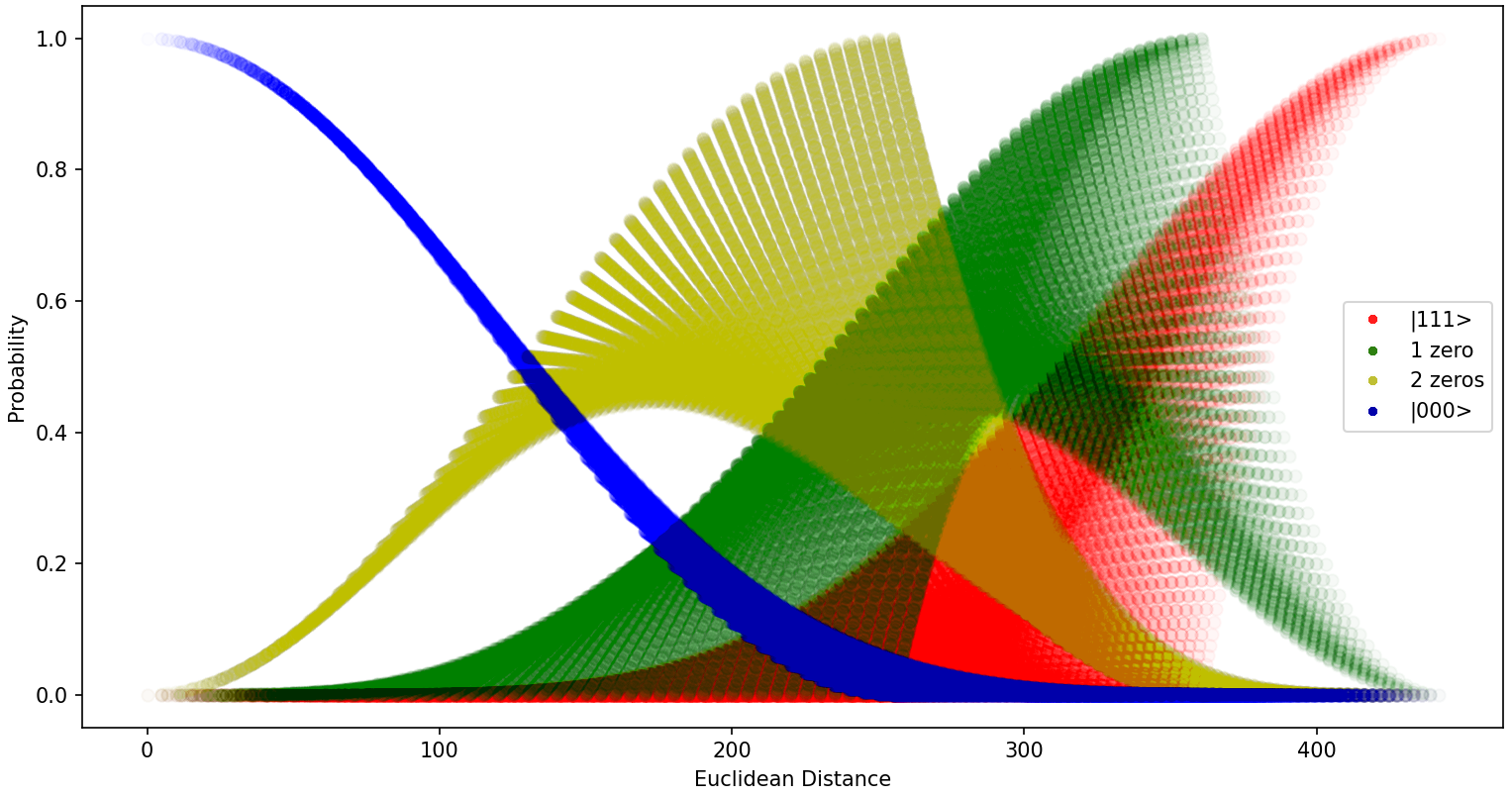}
	\caption{
		The probability of measuring a $\ket{000}$ outcome considering $\bm x$ from the full RGB space. Results are plotted along the Euclidean distance between RGB(0,0,0) and the input RGB vector.
		$\ket{001}, \ket{010}, \ket{100}$ grouped as ``2 zeros''. 
		$\ket{011}, \ket{101}, \ket{110}$ grouped as ``1 zero''.
		No $Q$ applied.
	}
	\label{fig:multiplot}
\end{figure}

    We implemented a 3-qubit model ($2^3$ basis states) relying on the IBMQ Qiskit framework \cite{ibm_quantum_experience_qiskit_2020}. To collect data about the probable outcomes, for each tested input $\bar{\bm x}$, we simulated $N=10^6$ measurements. In Tab. \ref{tab:colors} are reported some detailed operative examples, either in the canonical base or after a specific query, i.e., applying $Q(\bar{\bm x})$. In the first case, the basis states maintain a precise meaning, hence the corresponding colors are reported. In the second case, the Euclidean distance $d$ between the input and the target vector is added\footnote{
        Implemented using NumPy's  \href{https://numpy.org/doc/stable/reference/generated/numpy.linalg.norm.htm}{\texttt{norm}} function on the difference between the input and the target RGB vectors (not their normalized counterpart $\bm x$ and $\bar{\bm x}$).
    }. 
    To exhaustively explore the model's behavior through all the possible inputs, we sampled the RGB space with a sampling step of 5. Hence, we tested the model for $(255/5)^3=132651$ different inputs\footnote{
        The simulation took several days, but the average computational time of a single measurement simulation on a Core i5 10210U is $0.5 \cdot 10^{-2}$ seconds (tests available in \cite{lanza_quantum-robot_2020} via notebooks).
    }. We have not applied any $Q$ operator for these tests since the behavior would not change. Indeed, applying $Q$ changes only the basis states, not the behavior of the system in their regards. 
    
    Graphical visualizations of the behavior for three basis states and the $\ket{000}$ state relative to its distance to the input are reported in Fig. \ref{fig:cubestates} and Fig. \ref{fig:multiplot}, respectively.

\section{Experimental Insights}
\label{sec:discuss}

    The model behaves as expected. The confidence curve is shown in Fig. \ref{fig:multiplot} has a sinusoidal shape as observed in \cite{lanza_preliminary_2020}, which gives a nonlinear, yet definite, correspondence between stored information and measurements. As illustrated in Fig. \ref{fig:multiplot}, the more ``zeros'' the measured state has, the more is similar to $\ket{000}$. Even if Fig. \ref{fig:multiplot} refers to the canonical base, this behavior can be easily generalized using every state as a target applying $Q(\bar{\bm x})$ accordingly. We have to keep in mind that the $\ket{111}$ outcome is obtained only for ``extremely'' different input-target combinations, e.g., $\bar{\bm x}$ near $\ket{001}$ and $\bm x$ near $\ket{110}$. Targeting states in the whole RGB cube it is more likely to give readings which are, at most, a ``1 zero" measurement, as seen for the results reported in Tab. \ref{tab:colors}. 
    
    It is noteworthy that in our case study colors are just a graphical tool, not an actual concept. The answers are not to consider as precise statements about the color perceived by the system, rather a decision/classification process based on incomplete data. The probabilities indicate the degree of confidence the model has in answering a certain query in a certain way, based upon the previously collected knowledge. For this study, the knowledge relates only to a single instant, but for extended time windows ($\tau >1$ as in \cite{lanza_preliminary_2020}) this takes into account also the previous robot sensory history of the robot.

{
	\small
	\bibliographystyle{IEEEtran}
	\bibliography{0.myzoterolib.bib}

\begin{thebibliography}{10}
\providecommand{\url}[1]{#1}
\csname url@samestyle\endcsname
\providecommand{\newblock}{\relax}
\providecommand{\bibinfo}[2]{#2}
\providecommand{\BIBentrySTDinterwordspacing}{\spaceskip=0pt\relax}
\providecommand{\BIBentryALTinterwordstretchfactor}{4}
\providecommand{\BIBentryALTinterwordspacing}{\spaceskip=\fontdimen2\font plus
\BIBentryALTinterwordstretchfactor\fontdimen3\font minus
  \fontdimen4\font\relax}
\providecommand{\BIBforeignlanguage}[2]{{%
\expandafter\ifx\csname l@#1\endcsname\relax
\typeout{** WARNING: IEEEtran.bst: No hyphenation pattern has been}%
\typeout{** loaded for the language `#1'. Using the pattern for}%
\typeout{** the default language instead.}%
\else
\language=\csname l@#1\endcsname
\fi
#2}}
\providecommand{\BIBdecl}{\relax}
\BIBdecl

\bibitem{lanza_preliminary_2020}
\BIBentryALTinterwordspacing
D.~Lanza, P.~Solinas, and F.~Mastrogiovanni, ``A {Preliminary} {Study} for a
  {Quantum}-like {Robot} {Perception} {Model},'' \emph{arXiv:2006.02771
  [quant-ph]}, Jun. 2020, arXiv: 2006.02771. [Online]. Available:
  \url{http://arxiv.org/abs/2006.02771}
\BIBentrySTDinterwordspacing

\bibitem{petschnigg_quantum_2019}
\BIBentryALTinterwordspacing
C.~Petschnigg, M.~Brandstotter, H.~Pichler, M.~Hofbaur, and B.~Dieber,
  ``\BIBforeignlanguage{en}{Quantum {Computation} in {Robotic} {Science} and
  {Applications}},'' in \emph{\BIBforeignlanguage{en}{2019 {International}
  {Conference} on {Robotics} and {Automation} ({ICRA})}}.\hskip 1em plus 0.5em
  minus 0.4em\relax Montreal, QC, Canada: IEEE, May 2019, pp. 803--810.
  [Online]. Available: \url{https://ieeexplore.ieee.org/document/8793768/}
\BIBentrySTDinterwordspacing

\bibitem{amann_gestalt_1993}
\BIBentryALTinterwordspacing
A.~Amann, ``\BIBforeignlanguage{en}{The {Gestalt} problem in quantum theory:
  {Generation} of molecular shape by the environment},''
  \emph{\BIBforeignlanguage{en}{Synthese}}, vol.~97, no.~1, pp. 125--156, Oct.
  1993. [Online]. Available: \url{https://doi.org/10.1007/BF01255834}
\BIBentrySTDinterwordspacing

\bibitem{conte_testing_2008}
\BIBentryALTinterwordspacing
E.~Conte, ``\BIBforeignlanguage{en}{Testing {Quantum} {Consciousness}},''
  \emph{\BIBforeignlanguage{en}{NeuroQuantology}}, vol.~6, no.~2, Jun. 2008.
  [Online]. Available: \url{http://dx.doi.org/10.14704/nq.2008.6.2.167}
\BIBentrySTDinterwordspacing

\bibitem{conte_mental_2009}
\BIBentryALTinterwordspacing
E.~Conte, A.~Y. Khrennikov, O.~Todarello, A.~Federici, L.~Mendolicchio, and
  J.~P. Zbilut, ``\BIBforeignlanguage{en}{Mental {States} {Follow} {Quantum}
  {Mechanics} {During} {Perception} and {Cognition} of {Ambiguous}
  {Figures}},'' \emph{\BIBforeignlanguage{en}{Open Systems \& Information
  Dynamics}}, vol.~16, no.~01, pp. 85--100, Mar. 2009. [Online]. Available:
  \url{https://doi.org/10.1142/S1230161209000074}
\BIBentrySTDinterwordspacing

\bibitem{manousakis_quantum_2009}
\BIBentryALTinterwordspacing
E.~Manousakis, ``\BIBforeignlanguage{en}{Quantum formalism to describe
  binocular rivalry},'' \emph{\BIBforeignlanguage{en}{Biosystems}}, vol.~98,
  no.~2, pp. 57--66, Nov. 2009. [Online]. Available:
  \url{https://doi.org/10.1016/j.biosystems.2009.05.012}
\BIBentrySTDinterwordspacing

\bibitem{paraan_more_2014}
\BIBentryALTinterwordspacing
M.~R. Paraan, F.~Bakouie, and S.~Gharibzadeh, ``A more realistic quantum
  mechanical model of conscious perception during binocular rivalry,''
  \emph{Frontiers in Computational Neuroscience}, vol.~8, Feb. 2014. [Online].
  Available: \url{https://doi.org/10.3389/fncom.2014.00015}
\BIBentrySTDinterwordspacing

\bibitem{khrennikov_ubiquitous_2010}
\BIBentryALTinterwordspacing
A.~Y. Khrennikov, \emph{\BIBforeignlanguage{en}{Ubiquitous {Quantum}
  {Structure}}}.\hskip 1em plus 0.5em minus 0.4em\relax Berlin, Heidelberg:
  Springer Berlin Heidelberg, 2010. [Online]. Available:
  \url{http://link.springer.com/10.1007/978-3-642-05101-2}
\BIBentrySTDinterwordspacing

\bibitem{busemeyer_quantum_2012}
\BIBentryALTinterwordspacing
J.~R. Busemeyer and P.~D. Bruza, \emph{Quantum {Models} of {Cognition} and
  {Decision}}.\hskip 1em plus 0.5em minus 0.4em\relax Cambridge University
  Press, 2012. [Online]. Available:
  \url{https://doi.org/10.1017/CBO9780511997716}
\BIBentrySTDinterwordspacing

\bibitem{asano_quantum_2015}
\BIBentryALTinterwordspacing
M.~Asano, A.~Khrennikov, M.~Ohya, Y.~Tanaka, and I.~Yamato,
  \emph{\BIBforeignlanguage{en}{Quantum {Adaptivity} in {Biology}: {From}
  {Genetics} to {Cognition}}}.\hskip 1em plus 0.5em minus 0.4em\relax
  Dordrecht: Springer Netherlands, 2015. [Online]. Available:
  \url{http://link.springer.com/10.1007/978-94-017-9819-8}
\BIBentrySTDinterwordspacing

\bibitem{haven_palgrave_2017}
\BIBentryALTinterwordspacing
E.~Haven and A.~Khrennikov, Eds., \emph{\BIBforeignlanguage{en}{The {Palgrave}
  {Handbook} of {Quantum} {Models} in {Social} {Science}: {Applications} and
  {Grand} {Challenges}}}.\hskip 1em plus 0.5em minus 0.4em\relax London:
  Palgrave Macmillan UK, 2017. [Online]. Available:
  \url{http://link.springer.com/10.1057/978-1-137-49276-0}
\BIBentrySTDinterwordspacing

\bibitem{conte_algebraic_2018}
\BIBentryALTinterwordspacing
E.~Conte, \emph{\BIBforeignlanguage{English}{Algebraic {Quantum} {Theory} of
  {Consciousness}}}.\hskip 1em plus 0.5em minus 0.4em\relax Aracne, 2018.
  [Online]. Available:
  \url{http://www.aracneeditrice.it/aracneweb/index.php/pubblicazione.html?item=9788825517071}
\BIBentrySTDinterwordspacing

\bibitem{caves_quantum_2002}
\BIBentryALTinterwordspacing
C.~M. Caves, C.~A. Fuchs, and R.~Schack, ``Quantum probabilities as {Bayesian}
  probabilities,'' \emph{Physical Review A}, vol.~65, no.~2, p. 022305, Jan.
  2002. [Online]. Available: \url{http://doi.org/10.1103/PhysRevA.65.022305}
\BIBentrySTDinterwordspacing

\bibitem{busemeyer_what_2015}
\BIBentryALTinterwordspacing
J.~R. Busemeyer and Z.~Wang, ``What {Is} {Quantum} {Cognition}, and {How} {Is}
  {It} {Applied} to {Psychology}?'' \emph{Current Directions in Psychological
  Science}, vol.~24, no.~3, pp. 163--169, Jun. 2015, publisher: SAGE
  Publications Inc. [Online]. Available:
  \url{https://doi.org/10.1177/0963721414568663}
\BIBentrySTDinterwordspacing

\bibitem{nielsen_quantum_2010}
\BIBentryALTinterwordspacing
M.~A. Nielsen and I.~L. Chuang, \emph{\BIBforeignlanguage{en}{Quantum
  {Computation} and {Quantum} {Information}}}.\hskip 1em plus 0.5em minus
  0.4em\relax Cambridge University Press, Dec. 2010. [Online]. Available:
  \url{https://doi.org/10.1017/CBO9780511976667}
\BIBentrySTDinterwordspacing

\bibitem{hangos_state_2011}
\BIBentryALTinterwordspacing
K.~M. Hangos and L.~Ruppert, ``\BIBforeignlanguage{en}{State estimation methods
  using indirect measurements},'' in \emph{\BIBforeignlanguage{en}{Quantum
  {Probability} and {Related} {Topics}}}.\hskip 1em plus 0.5em minus
  0.4em\relax Santiago, Chile: World Scientific, Jan. 2011, pp. 163--180.
  [Online]. Available: \url{https://doi.org/10.1142/9789814338745_0009}
\BIBentrySTDinterwordspacing

\bibitem{ibm_quantum_experience_qiskit_2020}
\BIBentryALTinterwordspacing
\relax{IBM Quantum Experience}, ``Qiskit,'' https://github.com/Qiskit/qiskit.
  [Online]. Available: \url{https://github.com/Qiskit/qiskit}
\BIBentrySTDinterwordspacing

\bibitem{lanza_quantum-robot_2020}
\BIBentryALTinterwordspacing
D.~Lanza, ``Quantum-robot {Python} {Package},''
  https://github.com/Davidelanz/quantum-robot. [Online]. Available:
  \url{https://github.com/Davidelanz/quantum-robot}
\BIBentrySTDinterwordspacing

\end{thebibliography}
}

\end{document}